\documentclass{article}
\usepackage{spconf,amsmath,graphicx,amssymb}
\usepackage{color}
\usepackage{xcolor}
\usepackage{float}
\usepackage{subfigure}
\usepackage{url}
\usepackage{breakurl}

\usepackage{adjustbox}
\usepackage{array}

\newcolumntype{R}[2]{%
    >{\adjustbox{angle=#1,lap=\width-(#2)}\bgroup}%
    l%
    <{\egroup}%
}
\newcommand*\rot{\multicolumn{1}{R{-50}{1em}}}

\def\m#1{\mathsf{#1}}

\title{Towards Generalization of 3D Human Pose Estimation In The Wild}
\name{Renato Baptista\thanks{This work was funded by the National Research Fund (FNR), Luxembourg, under the projects C15/IS/10415355/3DACT/Bjorn Ottersten and AFRPPP/11806282. The authors are grateful to Artec3D, the volunteers, and to all present and former members of the $\mbox{CVI}^2$ group at SnT for participating in the data collection. The experiments presented in this paper were carried out using the HPC facilities of the University of Luxembourg.}, 
    Alexandre Saint, 
    Kassem Al Ismaeil, 
    Djamila Aouada}

\address{Interdisciplinary Center for Security, Reliability and Trust (SnT) \\
    University of Luxembourg, Luxembourg \\ 
{\tt\small \{renato.baptista, alexandre.saint, kassem.alismaeil, djamila.aouada\}@uni.lu}}

\begin{document}
%
\maketitle
\begin{abstract}

In this paper, we propose 3DBodyTex.Pose, a dataset that addresses the task of 3D human pose estimation in-the-wild. Generalization to in-the-wild images remains limited due to the lack of adequate datasets. Existent ones are usually collected in indoor controlled environments where motion capture systems are used to obtain the 3D ground-truth annotations of humans.
3DBodyTex.Pose offers high quality and rich data containing 405 different real subjects in various clothing and poses, and 81k image samples with ground-truth 2D and 3D pose annotations. 
These images are generated from 200 viewpoints among which 70 challenging extreme viewpoints. This data was created starting from high resolution textured 3D body scans and by incorporating various realistic backgrounds.
Retraining a state-of-the-art 3D pose estimation approach using data augmented with 3DBodyTex.Pose showed promising improvement in the overall performance, and a sensible decrease in the per joint position error when testing on challenging viewpoints. The 3DBodyTex.Pose is expected to offer the research community with new possibilities for generalizing 3D pose estimation from monocular in-the-wild images.


\end{abstract}
%

\vspace{-0.1cm}
\section{Introduction}
\label{sec:intro}
\vspace{-0.1cm}
In the past couple of years, human pose estimation has received a lot of attention from the computer vision community. The goal is to estimate the 2D or 3D position of the human body joints given an image containing a human subject. This has a significant number of applications such as sports, healthcare solutions~\cite{baptista2019home}, action recognition~\cite{baptista2019home,demisse2018pose}, and animations.

Due to the recent advances in Deep Neural Networks (DNN), the task of 2D human pose estimation a great improvement in results~\cite{chu2017multi,cao2018openpose,newell2016stacked}. This has been mostly achieved thanks to the availability of large-scale datasets containing 2D annotations of humans in many different conditions, e.g., in the wild~\cite{andriluka20142d}. In contrast, advances in the task of human pose estimation in 3D remains limited. The main reasons are the ambiguity of recovering the 3D information from a single image, in addition to the lack of large-scale datasets with 3D annotations of humans, specifically considering in-the-wild conditions. Existent datasets with 3D annotations are usually collected in a controlled environment using Motion Capture (MoCap) systems~\cite{h36m_pami}. Consequently, the variations in background and camera viewpoints remain limited. In addition, DNNs~\cite{zhou2016deep} trained on such datasets have difficulties generalizing well to environments where a lot of variation is present, e.g., scenarios in the wild. 

\begin{figure}[t]
    \centering
    \includegraphics[width=\columnwidth]{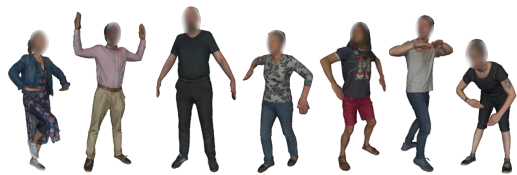}
    \caption{Examples of the 3D body scans used to generate in-the-wild images with 2D and 3D annotations of humans.}
    \label{fig:samples_dataset}
    \vspace{-0.35cm}
\end{figure}

Recently, many works focused on the challenging problem of 3D human pose estimation in the wild~\cite{zhou2017towards,yang20183d,pavllo20193d,rogez2019lcr}. These works differ significantly from each other 
but share an important aspect. They are usually evaluated on the same dataset that has been used for training. Thus, it is possible that these approaches have been over-optimized for specific datasets, leading to a lack of generalization. It becomes difficult to judge on the generalization, and more precisely for in-the-wild scenarios where variations coming from the background and camera viewpoints are always present.

In order to address the aforementioned challenge, this paper presents a new dataset referred to as \emph{3DBodyTex.Pose}. It is an original dataset generated from high-resolution textured 3D body scans, similar in quality to the ones contained in the 3DBodyTex dataset introduced in~\cite{Saint20183DBodyTexT3}. 3DBodyTex.Pose is dedicated to the task of human pose estimation. Synthetic scenes are generated with ground-truth information from real 3D body scans, with a large variation in subjects, clothing, and poses (see Fig.~\ref{fig:samples_dataset}). Realistic background is incorporated to the 3D environment. Finally, 2D images are generated from different camera viewpoints, including challenging ones, by virtually changing the camera location and orientation. We distinguish extreme viewpoints as the cases where the camera is, for example, placed on top of the subject. With the information contained in 3DBodyTex.Pose, it becomes possible to better generalize the problem of the 3D human pose estimation to in-the-wild images independently of the camera viewpoint as shown experimentally on a state-of-the-art 3D pose estimation approach~\cite{zhou2017towards}.
In summary, the contributions of this work are: \\
(1) 3DBodyTex.Pose, a synthetic dataset with 2D and 3D annotations of human poses considering in-the-wild images, with realistic background and standard to extreme camera viewpoints. This dataset will be publicly available for the research community.\\
(2) Increasing the robustness of 3D human pose estimation algorithms, specifically~\cite{zhou2017towards}, with respect to challenging camera viewpoints thanks to data augmentation with 3DBodyTex.Pose. 

The rest of the paper is organized as follows: Section~\ref{sec:relateddatasets} describes the related datasets for the 3D human pose estimation task. Section~\ref{sec:proposeddataet} provides details about the proposed 3DBodyTex.Pose dataset and how it addresses the challenges of in-the-wild images and extreme camera viewpoints. Then, Section~\ref{sec:experiments} shows the conducted experiments, and finally Section~\ref{sec:conclusion} concludes this work.

\vspace{-0.1cm}
\section{Related Datasets}
\label{sec:relateddatasets}
\vspace{-0.1cm}
Monocular 3D human pose estimation aims to estimate the 3D joint locations from the human present in the image independently of the environment of the scene. However, usually not all camera viewpoints are taken into consideration. Consequently, the 3D human body joints are not well estimated for the cases where the person is not fully visible or self-occluded. In order to use such images for training, labels for the position of the 2D human joints are needed as ground-truth information~\cite{andriluka20142d,johnson2010clustered}. Labeling such images from extreme camera viewpoints is an expensive and difficult task as it often requires manual annotation. To overcome this issue, MoCap systems can be used for precisely labeling the data. However, they are used in a controlled environment such as indoor scenarios. The Human3.6M dataset~\cite{h36m_pami} is widely used for the task of 3D human pose estimation and it falls under this scenario. It contains 3.6M frames with 2D and 3D annotations of humans from 4 different camera viewpoints. The HumanEva-I~\cite{sigal2010humaneva} and TotalCapture~\cite{trumble2017total} datasets are also captured in indoor environments. HumanEva-I contains 40k frames with 2D and 3D annotations from 7 different camera viewpoints. TotalCapture contains approximately 1.9M frames considering 8 camera locations where the 3D annotations of humans were obtained by fusing the MoCap with inertial measurement units. Also captured within a controlled environment, the authors of~\cite{mono-3dhp2017} proposed the MPII-INF-3DHP dataset for 3D human pose estimation which was recorded in a studio using a green screen background to allow automatic segmentation and augmentation. Consequently, the authors augment the data in terms of foreground and background, where the clothing color is changed on a pixel basis, and for the background, images sampled from the internet are used. Recently, von Marcard \textit{et al.}~\cite{vonMarcard2018} proposed a dataset with 3D pose in outdoor scenarios recorded with a moving camera. It contains more than 51k frames and 7 actors with a limited number of clothing style. \\
An alternative proposed with SURREAL~\cite{varol2017learning} and exploited in~\cite{pavlakos2018learning}, is to generate realistic ground-truth data synthetically. SURREAL places a parametric body model with varied pose and shape over a background image of a scene to simulate a monocular acquisition. Ground-truth 2D and 3D poses are known from the body model. To add realism, the body model is mapped with clothing texture. A drawback of this approach is that the body shape lacks details.\\
The 3DBodyTex dataset~\cite{Saint20183DBodyTexT3} contains static 3D body scans from people in close-fitting clothing, in varied poses and with ground-truth 3D pose. This dataset is not meant for the task of 3D human pose estimation. However, it is appealing for its realism: detailed shape and high-resolution texture information. It has been exploited for 3D human body fitting~\cite{Saint2019BodyfitrRA} and it could also be used to synthesize realistic monocular images from arbitrary viewpoints with ground-truth 2D and 3D poses. The main drawback of this dataset is the fact that it contains the same tight clothing with no variations.

\begin{table*}[t]
\vspace{-0.4cm}
\centering
\begin{tabular}{|l|ccccccc|}
\hline
                 & \rot{\textbf{\begin{tabular}[c]{@{}c@{}}3DBodyTex.\\ Pose (Ours)\end{tabular}}} & \rot{HumanEva-I} & \rot{Human3.6M}  &  \rot{\begin{tabular}[c]{@{}c@{}}MPII-INF-\\ -3DHP\end{tabular}} & \rot{TotalCapture} & \rot{3DPW} & \rot{SURREAL}  \\ 
\hline
\# of subjects            & 405                 & 4          & 11           & 8                       & 5            & 7                   & n/a           \\
\# of samples             & 81k                 & 40k        & $\sim$3.6M   & \textgreater{}1.3M      & $\sim$1.9M   & \textgreater{}51k   & $\sim$6.5M    \\
Ground-truth pose         & 2D+3D               & 2D+3D      & 2D+3D        & 3D                      & 3D           & 3D                  & 2D+3D         \\
Real people               & Yes                 & Yes        & Yes          & Yes                     & Yes          & Yes                 & No            \\
Background                & Indoor \& Outdoor   & Indoor     & Indoor       & Green Screen            & Indoor       & Outdoor             & Indoor     \\
Clothing                  & Realistic           & Realistic  & Realistic    & Realistic$^{(\star)}$  & No           & Limited             & No            \\
\# of total camera viewpoints   & 200                 & 7          & 4            & 14                      & 8            & n/a                 & n/a           \\ 
\# of challenging viewpoints & 70            & 0          & 0            & 3                       & 0            & n/a                 & n/a           \\ 
\hline
\end{tabular}
\caption{Comparison of datasets for the task of 3D human pose estimation. $(\star)$ indicates that clothing was synthetically added to the dataset.}
\label{tab:datasets_comparison}
\vspace{-0.1cm}
\end{table*}

\vspace{-0.1cm}
\section{Proposed 3DBodyTex.Pose Dataset}
\label{sec:proposeddataet}
\vspace{-0.1cm}
In contrast with 3DBodyTex, the new 3DBodyTex.Pose dataset contains 3D body scans that are captured from 405 subjects in their own regular clothes. From these 405 subjects, 204 are females and 201 are males. Having different clothing style from different people adds more variation to the dataset when considering in-the-wild scenarios. Fig.~\ref{fig:samples_dataset} shows a couple of examples of 3D body scans with different clothing. In this work, the goal is to use the 3D body scans to synthesize realistic monocular images from arbitrary camera viewpoints with its corresponding 2D and 3D ground-truth information for the task of 3D human pose estimation. The principal characteristics of 3DBodyTex.Pose are compared to state-of-the-art datasets in Table~\ref{tab:datasets_comparison}.

The 3DBodyTex.Pose dataset aims to address the challenges of in-the-wild images and the extreme camera viewpoints. Given that the only input is the set of 3D scans, we need to estimate the ground-truth 3D skeletons, to synthesize the monocular images from challenging viewpoints and to simulate an in-the-wild environment. These three stages are detailed below.
\begin{figure}[t]
    \vspace{-0.2cm}
    \centering
    \includegraphics[width=\columnwidth]{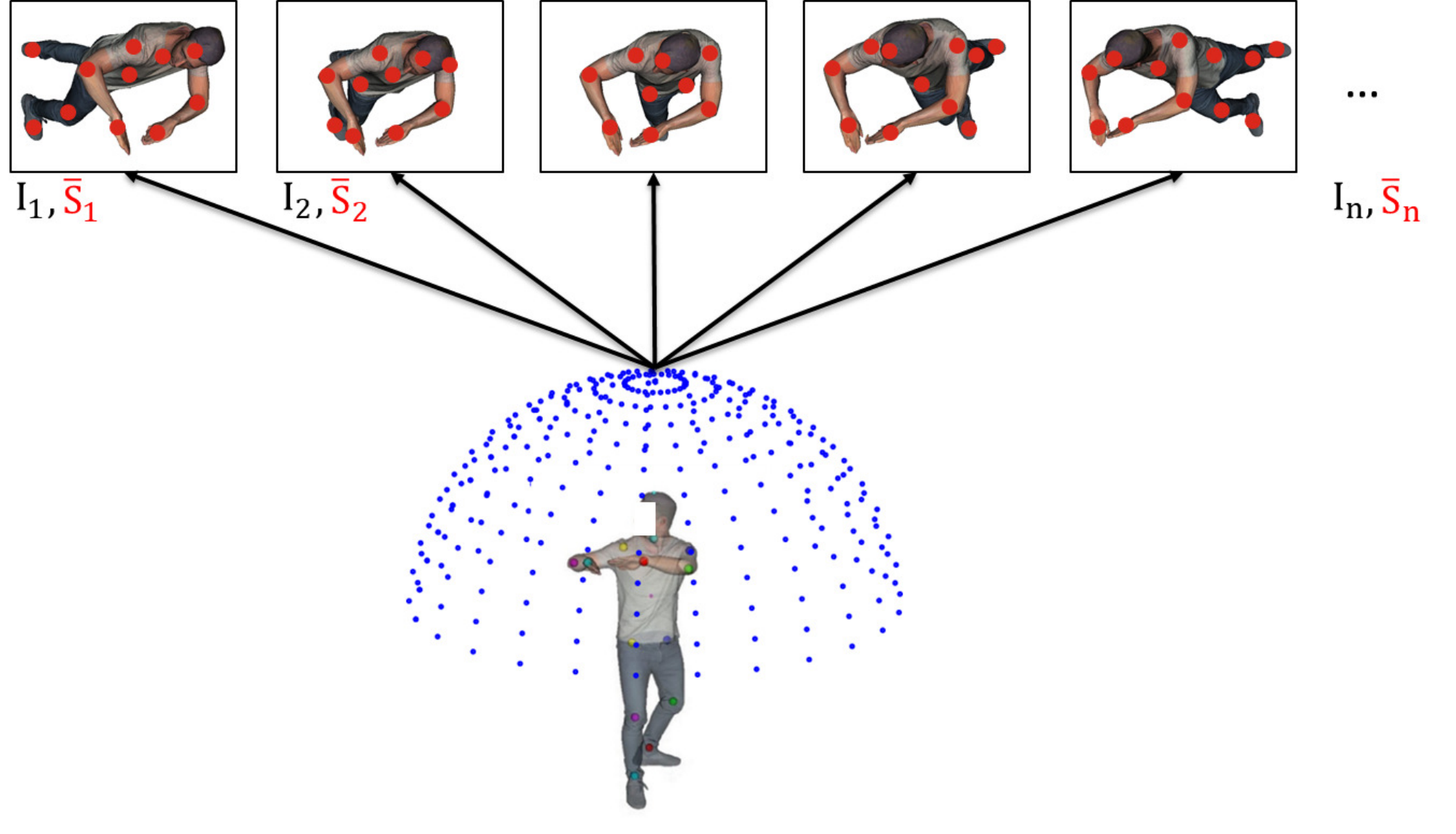}
    \caption{Extreme camera viewpoints images (top row) from a single 3D body scan. The blue dots represent the camera locations for each camera viewpoint. Better visualized in color.}
    \label{fig:camera_extreme_viewpoints}
    \vspace{-0.2cm}
\end{figure}

\begin{figure}[t]
    \vspace{-0.5cm}
	\centering 
    \subfigure[]{{\includegraphics[width=0.51\columnwidth]{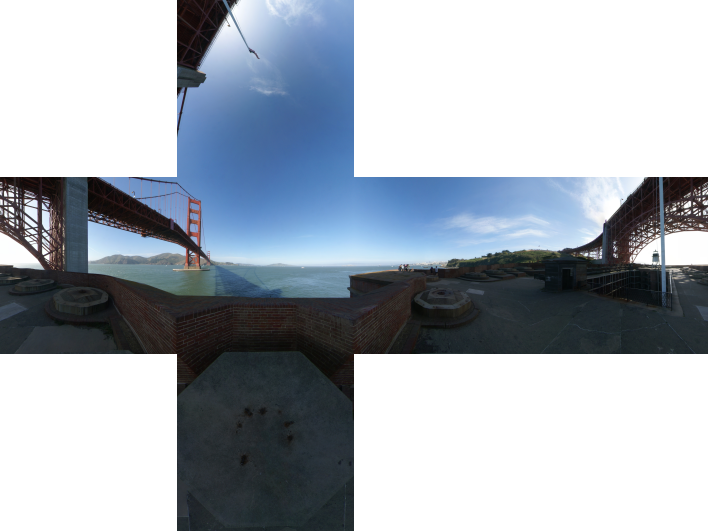}}\label{fig:cube_mapping}}
    \; \; \;
    \subfigure[]{{\includegraphics[width=0.40\columnwidth]{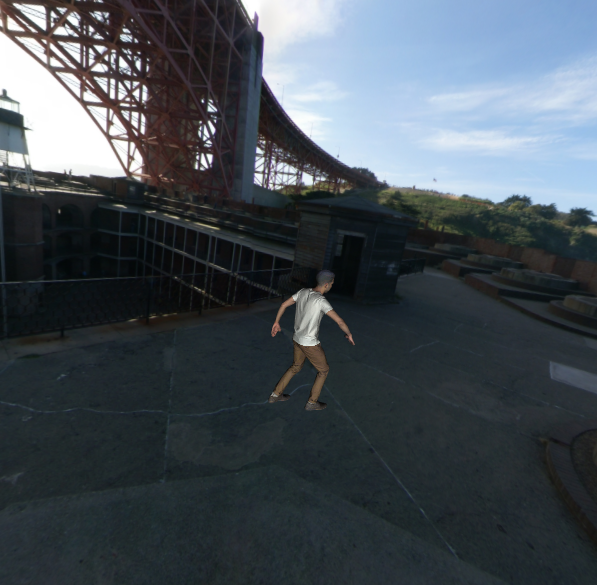}}\label{fig:cube_3d}}
    \vspace{-0.3cm}
    \caption{(a) Example of an unfolded cube projection of a 3D environment (extracted from~\cite{humus-cube}). (b) Example of a 3D body scan added to the 3D environment of a realistic scene.}
    \label{fig:all_cube_3d}
    \vspace{-0.2cm}
\end{figure}

\noindent{\bf{Ground-truth 3D joints.}} To estimate the ground-truth 3D skeleton, we follow the automatic approach of 3DBodyTex~\cite{Saint20183DBodyTexT3} where body landmarks are first detected in 2D views before being robustly aggregated into 3D positions. Hence, for every 3D scan we have the corresponding 3D positions of the human body joints that is henceforth used as the ground-truth 3D skeleton.\\ 
\noindent{\bf{Challenging viewpoints.}} We propose to change the location and orientation of the camera in order to create monocular images where also extreme viewpoints are considered, see Fig.~\ref{fig:camera_extreme_viewpoints}. Considering a 3D body scan~$\m P \in \mathbb{R}^{3 \times K}$, where~$K$ is the number of vertices of the mesh, the 3D skeleton with~$J$ joints~$\m S \in \mathbb{R}^{3 \times J}$, and also the homogeneous projection matrix~$\m M_n$ for the camera position~$n$, we can back-project the 3D skeleton into the image~$\m I_n$ by
\begin{equation}
    \vspace{-0.1cm}
    \centering
    \bar{\m S}_n = \m M_n \cdot \m S,
    \label{eq:reproject}
    \vspace{-0.1cm}
\end{equation}
where~$\bar{\m S}_n \in \mathbb{R}^{3 \times J}$ represents the homogeneous coordinates of the projected 3D skeleton into the image plane~$\m I_n$, corresponding to a 2D skeleton. In this way, we are able to generate all possible camera viewpoints around the subject and easily obtain the corresponding 2D skeleton. In summary, each element of the 3DBodyTex.Pose is composed of image~$\m I_n$, 2D skeleton~$\bar{\m S}_n$, and 3D skeleton~$\m S$ in the camera coordinate system.\\
\noindent{\bf{In-the-wild environment.}} In order to address the challenge of the in-the-wild images with ground-truth information for the task of 3D human pose estimation, we further propose to embed the 3D scan in an environment with cube mapping~\cite{greene1986environment} which in turns adds a realistic background variation to the dataset. An example texture cube is shown in Figure~\ref{fig:cube_mapping}. The six faces are mapped to a cube surrounding the scene with the 3D body scan at the center, see Figure~\ref{fig:cube_3d}. Realistic textures cubes are obtained from~\cite{humus-cube}.

To have variation in the data, for each image, we randomly draw a texture cube, a camera viewpoint and a 3D scan. The proposed 3DBodyTex.Pose dataset provides reliable ground-truth 2D and 3D annotations with realistic and varied in-the-wild images while considering arbitrary camera viewpoints. Moreover, it offers a relatively high number of subjects in comparison with state-of-the-art 3D pose datasets, refer to Table~\ref{tab:datasets_comparison}. It also offers richer body details in terms of clothing, shape, and the realistic texture.

\vspace{-0.1cm}
\section{Experimental Evaluation}
\label{sec:experiments}
\vspace{-0.1cm}
In what follows, we use the approach proposed by Zhou \textit{et al.}~\cite{zhou2017towards} to showcase the impact of the 3DBodyTex.Pose dataset in improving the performance of 3D pose estimation in the wild. We note that, in a similar fashion, 3DBodyTex.Pose can be used to enhance any other existent approach. Our goal is to share this new dataset with the research community and encourage (re-)evaluating and (re-)training existent and new 3D pose estimation approaches especially considering in-the-wild scenarios with a special focus on extreme viewpoints.

\vspace{-0.1cm}
\subsection{Baseline 3D Pose Estimation Approach}
\vspace{-0.1cm}
The work in~\cite{zhou2017towards} aims to estimate 3D human poses in the wild. For that, the authors proposed to couple together in-the-wild images with 2D annotations with indoor images with 3D annotations in an end-to-end framework. The authors provided the code for both training and testing the network. \\
The network proposed in~\cite{zhou2017towards} consists of two different modules: (1) 2D pose estimation module; and (2) depth regression module. In the first module, the goal is to predict a set of~$J$ heat maps by minimizing the~$L^2$ distance between the predicted and the ground-truth heat maps where only images with 2D annotations were used (MPII dataset~\cite{andriluka20142d}). Secondly, the depth regression module learns to predict the depth between the camera and the image plane by using the images where 3D annotations are provided (Human3.6M dataset~\cite{h36m_pami}). Also within the second module, the authors proposed a geometric constraint which serves as a regularization for depth prediction when the 3D annotations are not available. At the end, the network is built in a way that both modules are trained together. 



\vspace{-0.1cm}
\subsection{Data Augmentation with 3DBodyTex.Pose}
\vspace{-0.1cm}
We propose to retrain the network presented in~\cite{zhou2017towards} by adding the 3DBodyTex.Pose data to the training set originally used in~\cite{zhou2017towards}. Specifically, 60k additional RGB images from 3DBodyTex.Pose and their corresponding 2D skeletons were used to increase the variation coming from realistic background and camera viewpoints. \\
We first follow the same evaluation protocol as in~\cite{zhou2017towards} by testing on the Human3.6M dataset~\cite{h36m_pami}, and using the Mean Per Joint Position Error (MPJPE) in millimeters (mm) as an evaluation metric between 3D skeletons.
Table~\ref{tab:h36m_avg_results} shows the results of retraining~\cite{zhou2017towards} by augmenting with 3DBodyTex.Pose (\textbf{Zhou \textit{et al.}~\cite{zhou2017towards} ++}) along with other reported state-of-the-art results as a reference.
 Without using 3DBodyTex.Pose, the average error between the estimated 3D skeleton and the ground-truth annotation is 64.9 mm, and when retrained with the addition of our proposed dataset, the error decreases to 61.3 mm. This result is a very promising step towards the generalization of 3D human pose estimation for in-the-wild images. Despite the fact that testing in Table~\ref{tab:h36m_avg_results} is on Human3.6M (indoor scenes only), retraining with 3DBodyTex.Pose helps bring the performance of~\cite{zhou2017towards} closer to the top performing approaches~\cite{rogez2019lcr,yang20183d} and even beating others, i.e.,~\cite{martinez2017simple}. 
 

\begin{table}[t]
\centering
\begin{tabular}{l|c} 
\hline
Methods & Average (mm)  \\ 
\hline
Zhou~\textit{et al.}~\cite{zhou2017towards}                       &  64.9  
\\
\textbf{Zhou \textit{et al.}~\cite{zhou2017towards} ++ (Ours)}         &  \textbf{61.3}  
\\
\hline
\hline
Martinez~\textit{et al.}~\cite{martinez2017simple}                &  62.9   \\
Rogez~\textit{et al.}~\cite{rogez2019lcr}                         &  61.2   \\
Yang~\textit{et al.}~\cite{yang20183d}                            &  \underline{58.6}   \\
\end{tabular}
\caption{Quantitative results of the MPJPE in millimeters on the Human3.6M dataset following the same protocol as in~\cite{zhou2017towards}. The average column represents the average error value of all actions in the validation set.}
\label{tab:h36m_avg_results}
\end{table}

\noindent As one of the aims of this paper is to mitigate the effect of challenging camera viewpoints, we tested the performance of~\cite{zhou2017towards} on a new testing set containing challenging viewpoints only. These were selected from the 3DBodyTex.Pose dataset and reserved for testing only\footnote{Never seen during training.}. Table~\ref{tab:extreme_vp} shows that adding the 3DBodyTex.Pose to the training set in the network of~\cite{zhou2017towards} performs better when testing with challenging viewpoints only. Note that the relative high values of the errors, as compared to Table~\ref{tab:h36m_avg_results}, are due to the fact that the depth regression module is learned with the 3D ground-truth poses of the Human3.6M dataset only.

\begin{table}[t]
\vspace{-0.1cm}
\centering
\begin{tabular}{l|c}
\hline
Methods                                                                 & Average (mm) \\ \hline
Zhou~\textit{et al.}~\cite{zhou2017towards}                             & 292          \\
\textbf{Zhou \textit{et al.}~\cite{zhou2017towards} ++ (Ours)}   & \textbf{267}         
\end{tabular}
\caption{Results of the MPJPE while testing on challenging camera viewpoints only.}
\label{tab:extreme_vp}
\vspace{-0.2cm}
\end{table}



\vspace{-0.2cm}
\section{Conclusion}
\label{sec:conclusion}
\vspace{-0.1cm}
This paper introduced the 3DBodyTex.Pose dataset as a new original dataset to support the research community in designing robust approaches for 3D human pose estimation in the wild, independently of the camera viewpoint. It contains synthetic but realistic monocular images with 2D and 3D human pose annotations, generated from diverse and high-quality textured 3D body scans. The potential of this dataset was demonstrated by retraining a state-of-the-art 3D human pose estimation approach. There is a significant improvement in performance when augmented with 3DBodyTex.Pose. This opens the door to the generalization of 3D human pose estimation to in-the-wild images. As future work, we intend to increase the size of the dataset covering more camera viewpoints and realistic backgrounds, 
and by adding different scaling factors with respect to the camera location in order to increase the generalization over the depth variation.


\bibliographystyle{IEEEbib}
\bibliography{refs}

\begin{thebibliography}{10}

\bibitem{baptista2019home}
Renato Baptista, Enjie Ghorbel, Abd El~Rahman Shabayek, Florent Moissenet,
  Djamila Aouada, Alice Douchet, Mathilde Andr{\'e}, Julien Pager, and
  St{\'e}phane Bouilland,
\newblock ``Home self-training: Visual feedback for assisting physical activity
  for stroke survivors,''
\newblock {\em CMPB}, 2019.

\bibitem{demisse2018pose}
Girum~G Demisse, Konstantinos Papadopoulos, Djamila Aouada, and Bjorn
  Ottersten,
\newblock ``Pose encoding for robust skeleton-based action recognition,''
\newblock in {\em CVPRW}, 2018.

\bibitem{chu2017multi}
Xiao Chu, Wei Yang, Wanli Ouyang, Cheng Ma, Alan~L Yuille, and Xiaogang Wang,
\newblock ``Multi-context attention for human pose estimation,''
\newblock in {\em CVPR}, 2017.

\bibitem{cao2018openpose}
Zhe Cao, Gines Hidalgo, Tomas Simon, Shih-En Wei, and Yaser Sheikh,
\newblock ``Openpose: realtime multi-person 2d pose estimation using part
  affinity fields,''
\newblock {\em arXiv:1812.08008}, 2018.

\bibitem{newell2016stacked}
Alejandro Newell, Kaiyu Yang, and Jia Deng,
\newblock ``Stacked hourglass networks for human pose estimation,''
\newblock in {\em ECCV}, 2016.

\bibitem{andriluka20142d}
Mykhaylo Andriluka, Leonid Pishchulin, Peter Gehler, and Bernt Schiele,
\newblock ``2d human pose estimation: New benchmark and state of the art
  analysis,''
\newblock in {\em CVPR}, 2014.

\bibitem{h36m_pami}
Catalin Ionescu, Dragos Papava, Vlad Olaru, and Cristian Sminchisescu,
\newblock ``Human3.6m: Large scale datasets and predictive methods for 3d human
  sensing in natural environments,''
\newblock {\em IEEE TPAMI}, vol. 36, no. 7, pp. 1325--1339, jul 2014.

\bibitem{zhou2016deep}
Xingyi Zhou, Xiao Sun, Wei Zhang, Shuang Liang, and Yichen Wei,
\newblock ``Deep kinematic pose regression,''
\newblock in {\em ECCV}, 2016.

\bibitem{zhou2017towards}
Xingyi Zhou, Qixing Huang, Xiao Sun, Xiangyang Xue, and Yichen Wei,
\newblock ``Towards 3d human pose estimation in the wild: a weakly-supervised
  approach,''
\newblock in {\em ICCV}, 2017.

\bibitem{yang20183d}
Wei Yang, Wanli Ouyang, Xiaolong Wang, Jimmy Ren, Hongsheng Li, and Xiaogang
  Wang,
\newblock ``3d human pose estimation in the wild by adversarial learning,''
\newblock in {\em CVPR}, 2018.

\bibitem{pavllo20193d}
Dario Pavllo, Christoph Feichtenhofer, David Grangier, and Michael Auli,
\newblock ``3d human pose estimation in video with temporal convolutions and
  semi-supervised training,''
\newblock in {\em CVPR}, 2019.

\bibitem{rogez2019lcr}
Gr{\'e}gory Rogez, Philippe Weinzaepfel, and Cordelia Schmid,
\newblock ``Lcr-net++: Multi-person 2d and 3d pose detection in natural
  images,''
\newblock {\em IEEE TPAMI}, 2019.

\bibitem{Saint20183DBodyTexT3}
Alexandre Saint, Eman Ahmed, Abd El~Rahman Shabayek, Kseniya Cherenkova, Gleb
  Gusev, Djamila Aouada, and Bj{\"o}rn Ottersten,
\newblock ``3dbodytex: Textured 3d body dataset,''
\newblock {\em 3DV}, 2018.

\bibitem{johnson2010clustered}
Sam Johnson and Mark Everingham,
\newblock ``Clustered pose and nonlinear appearance models for human pose
  estimation.,''
\newblock in {\em BMVC}. Citeseer, 2010, vol.~2, p.~5.

\bibitem{sigal2010humaneva}
Leonid Sigal, Alexandru~O Balan, and Michael~J Black,
\newblock ``Humaneva: Synchronized video and motion capture dataset and
  baseline algorithm for evaluation of articulated human motion,''
\newblock {\em IJCV}, vol. 87, no. 1-2, pp. 4, 2010.

\bibitem{trumble2017total}
Matthew Trumble, Andrew Gilbert, Charles Malleson, Adrian Hilton, and John
  Collomosse,
\newblock ``Total capture: 3d human pose estimation fusing video and inertial
  sensors.,''
\newblock in {\em BMVC}, 2017, vol.~2, p.~3.

\bibitem{mono-3dhp2017}
Dushyant Mehta, Helge Rhodin, Dan Casas, Pascal Fua, Oleksandr Sotnychenko,
  Weipeng Xu, and Christian Theobalt,
\newblock ``Monocular 3d human pose estimation in the wild using improved cnn
  supervision,''
\newblock in {\em 3DV}, 2017.

\bibitem{vonMarcard2018}
Timo von Marcard, Roberto Henschel, Michael Black, Bodo Rosenhahn, and Gerard
  Pons-Moll,
\newblock ``Recovering accurate 3d human pose in the wild using imus and a
  moving camera,''
\newblock in {\em ECCV}, sep 2018.

\bibitem{varol2017learning}
Gul Varol, Javier Romero, Xavier Martin, Naureen Mahmood, Michael~J Black, Ivan
  Laptev, and Cordelia Schmid,
\newblock ``Learning from synthetic humans,''
\newblock in {\em CVPR}, 2017.

\bibitem{pavlakos2018learning}
Georgios Pavlakos, Luyang Zhu, Xiaowei Zhou, and Kostas Daniilidis,
\newblock ``Learning to estimate 3d human pose and shape from a single color
  image,''
\newblock in {\em CVPR}, 2018.

\bibitem{Saint2019BodyfitrRA}
Alexandre Saint, Abd El~Rahman Shabayek, Kseniya Cherenkova, Gleb Gusev,
  Djamila Aouada, and Bj{\"o}rn Ottersten,
\newblock ``Bodyfitr: Robust automatic 3d human body fitting,''
\newblock {\em ICIP}, 2019.

\bibitem{greene1986environment}
Ned Greene,
\newblock ``Environment mapping and other applications of world projections,''
\newblock {\em IEEE CG\&A}, vol. 6, no. 11, pp. 21--29, 1986.

\bibitem{humus-cube}
``{Humus Cubemap},'' \url{http://www.humus.name/index.php?page=Textures},
\newblock Accessed: 2020-01-29.

\bibitem{martinez2017simple}
Julieta Martinez, Rayat Hossain, Javier Romero, and James~J Little,
\newblock ``A simple yet effective baseline for 3d human pose estimation,''
\newblock in {\em ICCV}, 2017.

\end{thebibliography}


\end{document}